\documentclass{ws-ijprai}
\begin{document}

\markboth{H. Zhou \& Y. Li}{Using Deep Neural Network Approximate Bayesian Network}

%
\catchline{}{}{}{}{}
%

\title{Using Deep Neural Network Approximate Bayesian Network}

\author{Jie Jia}
\address{Taiyuan University of Technology, Taiyuan, China\,
\email{jiajie1921@link.tyut.edu.cn}}
\author{Honggang Zhou}

\address{School of Computer Science and Engineering, Beihang University, Beijing, China\,
\email{zhg@buaa.edu.cn}}

\author{Yunchun Li}

\address{School of Computer Science and Engineering, Beihang University, Beijing, China\,\\
lych@buaa.edu.cn}

\maketitle


\begin{abstract}
We present a new method to approximate posterior probabilities of Bayesian Network using Deep Neural Network. Experiment results on several public Bayesian Network datasets shows that Deep Neural Network is capable of learning joint probability distribution of Bayesian Network by learning from a few observation and posterior probability distribution pairs with high accuracy. Compared with traditional approximate method likelihood weighting sampling algorithm, our method is much faster and gains higher accuracy in medium sized Bayesian Network. Another advantage of our method is that our method can be parallelled much easier in GPU without extra effort. We also explored the connection between the accuracy of our model and the number of training examples. The result shows that our model saturate as the number of training examples grow and we don't need many training examples to get reasonably good result. Another contribution of our work is that we have shown discriminative model like Deep Neural Network can approximate generative model like Bayesian Network.
\end{abstract}


\section{Introduction}
Bayesian Network(BN) is a generative model representing joint probabilities of random variables by decomposing them into prior probability distribution and conditional probability distribution which can be represented by a directed graphical model. The joint probability is:
 \begin{equation}
 P(x_1,...,x_n)=\prod_{i=1}^nP(x_i|Parents(x_i))
 \end{equation}
 $x_i$ is random variable in BN, $Parents(x_i)$ is variables $x_i$ conditioned to and correspond to the parent nodes of $x_i$ in \emph{directed acyclic graph(DAG)} of BN. BN has been successfully used in various domains. D.J.Spiegelhalter etc. used BN to develop an expert system diagnosing \"blue\" baby disease and compared BN expert system with traditional logic based expert system\cite{spiegelhalter1993bayesian}. R.B.Cowell etc. summarize the use of BN in expert system and inference method widely used in it\cite{probabilitisexpert}. N. Friedman etc. applied BN to find interactions between genes\cite{computationalbiology}. Cai etc. developed an ground-source heat fault diagnose system using BN and obtained better performance than traditional method\cite{heatpump}. Also BN has been applied in risk analysis of Maritime Transport System by P. Trucco etc.\cite{riskanalysis}. BN's advantages include:
\begin{itemlist}
 \item can represent any kind of probability distribution and mutual dependence,
 \item can incorporate prior distribution (prior kownledge),
 \item can reason from any observations to any latent variables,
 \item can cope with incomplete and uncertain data\cite{bayesianandneuralnetwork},
 \item not easily over overfit (chapter 5 in Ref.8).
\end{itemlist}

However, BN's exact inference is a NP-hard problem for a general graphical model which has been proved by G. Cooper\cite{cooper1990}. There are also many approximate inference methods including \emph{Markov Chain Monte Carlo (MCMC)\cite{mcmc}, Loopy Belief Propagation\cite{loopybeliefpropagation}, Logic Sampling\cite{logicsampling} }etc. P. Dagum's work shows that even approximate inference is also NP-hard\cite{approximatingnphard}. This is the main reason preventing BN from accommodating more nodes.

Inspired by the success of Deep Neural Network (DNN) recent years in computer vision\cite{fasterrcnn,inceptionresnet}, natural language process\cite{nlppaper} etc., we have tried to connect DNNs and BN. DNN has powerful non-linear fitting and automatic feature extraction ability. The challenge is that DNN is a discriminant model and BN is a generative model. We solved this problem by making DNN predict the logarithmic probability of every possible value of stochastic variables in BN from various observations as input. We have only experimented on discrete BN so far. Dataset for DNN is obtained by randomly choosing a group of observations and correspondingly calculate posterior probability distribution for each stochastic variable using exact inference algorithm (junction tree algorithm\cite{junctiontreealgorithm} in our experiments). Then we separated the dataset into training set and testing set.

Our experiments show that after learning from a few observations, DNN can actually "understand" the inner structure of BN and perform inference. Here "understand" means that DNN can inference posterior probability distribution from observations that don't exit in training set. We are not the first to try to connect Neural Network and BN. A. Stassopoulou etc. found a correspondence between Neural Network and BN by deriving a closed-form solution so that the outputs of the two networks are similar in the least square error sense\cite{neuralnetworkandbn}. But they use hand-crafted structure and parameters which is not suitable for large network. Formally speaking, our model can be written in the following formula:
\begin{equation}P(V_i=V_{ij}|O)=\frac{exp(f_{ij}(O;\theta))}{\sum_jexp(f_{ij}(O;\theta))}\end{equation} where $V_{ij}$ is a possible value for random variable $V_i$ in BN, O represents for observation, ${\theta}$ represents for the parameters of DNN, $f_{ij}$ is ij-th output of DNN.

We haven't change the fact that inference of BN is NP-hard because we need exact inference to generate training dataset for DNN. But once trained, DNN can perform inference in polynomial time and can be easily parallelled using GPU. In fact, we transform the difficulty of BN inference to training period of DNN. Our main contributions are two-fold:
\begin{itemlist}
\item we propose a polynomial time approximate inference method for BN,
\item we demonstrate that DNN can learn inner structure of generative BN.
\end{itemlist}

\section{Related Work}

BN inference is to calculate the posterior probability distribution of all variables $P(V_i|O)$ given observation $O$. There are two classes of inference methods in BN: exact inference and approximate inference. J. H. Kim etc. developed an efficient message propagation exact inference algorithm which run in polynomial time but it applies only to polytrees\cite{polytree}. Another prevalent exact inference algorithm is \emph{junction tree algorithm} proposed by SL. Lauritzen etc.\cite{junctiontreealgorithm} which can calculate posterior probability distribution for all variables in one pass and can apply to any BN structure but it's runing time is exponential. Junction tree algorithm transform DAG of BN into a clique tree by moralize and triangularization and perform message propagation over the junction tree. Another exact inference algorithm is \emph{Variable Elimination(VE)} developed by N. L. Zhang etc.\cite{vealgorithm} which calculate a variable's posterior probability distribution by eliminate variable one by one by summing them out. As for approximate inference algorithm, there are mainly two kinds of algorithms: importance sampling and MCMC sampling. Importance sampling draws samples independent of each other, algorithms include forward sampling like logic sampling\cite{logicsampling}, likelihood weighting\cite{likelihoodweighting}, adaptive importance sampling\cite{adaptiveimportancesampling} and backward sampling\cite{backwardsampling}. MCMC sampling draw new sample dependent of previous sample, prevalent algorithms include Gibbs sampling\cite{fastgibbs} and Metropolis sampling. Except for developing new algorithms, there are some researches trying to parallelize these inference methods to reduce inference time. Zheng. L proposed a parallel junction tree algorithm running on GPU\cite{paralleljuctiontree}. Haoyu. C proposed an more efficient parallel Gibbs sampling algorithm running on GPU using state replication (State Augmented Marginal Estimation)\cite{fastgibbs}.

\section{Methodology}

\begin{figure}[bh]
\centerline{\includegraphics[width=9cm]{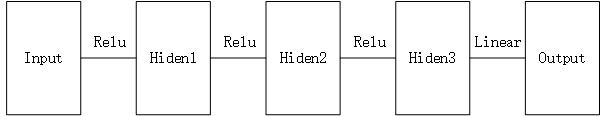}}
\vspace*{8pt}
\caption{DNN architecture used in the experiment. Input is a vector representing for observations. The dimension of input vector is $\sum_i|V_i|$, and $|V_i|$ is the number of possible values of random variable $V_i$, the dimension of output vector is the same as input vector.}
\end{figure}

We use one-hot vector for input (observation) embedding. The total number of input nodes $|\bm{X}|=\sum_i|V_i|$, where $|V_i|$ is the number of possible values of random variable $V_i$. Every possible value of of random variables has a corresponding input node. We set $\bm{X}_i$ 1 if the corresponding value of BN is observed. There are several fully-connected hidden layers after input layer. The activation function of those hidden layers is rectified linear function (relu):
\begin{equation}
relu(x)=max(0,x)
\end{equation}
Relu is widely used in modern DNN\cite{relu} because it is non-linear and can properly propagate gradient from upper layer. The output layer has the same dimension as input layer but it represent for logits of probabilities. The probability of random variable i take value j is:
\begin{equation}
\begin{aligned}
P(V_i=V_{ij})=&multi-softmax(logits_{ij})\\
=&\frac{exp(logits_{ij})}{\sum_jexp(logits_{ij})}
\end{aligned}
\end{equation}
Multi-softmax means there are multiple random variables and multiple values for each random variable. The loss function of DNN is:
\begin{equation}
\begin{aligned}
\mathcal{L}=&\sum_i^I\sum_j^{J_i}crossentropy(y_{ij},P(V_i=V_{ij}))+l_2norm(\theta)\\
=&\sum_i^I\sum_j^{J_i}-y_{ij}exp(P(V_i=V_{ij}))+\frac{\lambda}{I}\sum\theta\\
\end{aligned}
\end{equation}
where $y_{ij}$ is the truth probability for random variable $V_i$ taking value $V_{ij}$, I is the number of random variables, $J_i$ is the number possible values for variable $V_i$. The reason why we don't make DNN predict probability directly is that the output is real number but the probability is in range [0,1] and sum to 1.

We use multiple metrics to evaluate our model. The first metric we use is average Kullback-Leibler divergence (K-L distance) for all testing examples:
\begin{equation}
D_{avg}=\frac1N\sum_n^N\sum_i^{I}\sum_j^{J_i}y_{ij}^nlog\left(\frac{y_{ij}^n}{P^n(V_i=V_{ij})}\right)
\end{equation}
 where $N$ is the total number of testing examples. This metric used to calculated the divergence between target distribution $y_{ij}^n$ and calculated distribution $P^n(V_i=V_{ij})$. This metric has also been used to evaluate BN approximate inference algorithm by H.Chen\cite{fastgibbs}. One drawback of K-L metric is that it cannot show how many samples are predicted well (given a threshold). So We use multi-threshold accuracy (MTA) to solve this problem:
\begin{equation}
MTA=\frac1N\sum_n^N\frac1I\sum_i^I\frac1J_i\sum_j^{J_i}\bm1\{|y^n_{ij}-P^n(V_i=V_{ij})|<threshold\}
\end{equation}
$\bm1\{|y_{ij}-P(V_i=V_{ij})|<threshold\}$ is 1 when condition is satisfied and 0 otherwise. This metric generally indicates how many times our model make "serious"  mistakes when predicting probabilities.
\section{Experiment}

We have experimented on several public BN datasets to verify our model including small sized dataset Asia\cite{asia}, Survey\cite{survey} and medium sized dataset Alarm\cite{alarm}, Insurance\cite{insurance}. We didn't experiment on larger dataset because training data for DNN is hard to obtain. Larger BN may take minutes or even hours to perform an inference. We mainly compare our model to Likelihood Weighting Sampling (LWS) method\cite{likelihoodweighting}. LWS method is a melioration to forward sampling which draw samples following the influence arrow of BN. When met with samples inconsistent with observed random variables, LWS weights the sample by the likelihood of evidence conditioned on the sample. Probability for random variable $V_i$ taking value $V_{ij}$ is:
\begin{equation}
P(V_i=V_{ij})=\frac1N\sum_n^Nweight^n\bm1\{sample_i^n=V_{ij}\}
\end{equation}
We use Tensorflow to build our DNN model and generate dataset using junction tree algorithm. We run LWS and DNN on the same machine which is equipped with Intel core i5-3230M. We didn't use GPU to accelerate DNN for convenience of comparing running time. The L2 normalization coefficient we use is 0.005 across all datasets. Batchsize is 32 and learning rate is 0.0001. The optimizer we use is Momentum Optimizer\cite{momentum} with momentum rate 0.9. Momentum update weights as follow:
\begin{equation}
\begin{aligned}
&v_{t+1}=\mu v_t-\epsilon \nabla f(\theta_t) \\
&\theta_{t+1}=\theta+v_{t+1}
\end{aligned}
\end{equation}
Where $\epsilon>0$ is the learning rate, $\mu \in [0,1]$ is the momentum coefficient, $\nabla f(\theta_t)$ is the gradient at $\theta_t$. Here is some tricks we used during training:
 \begin{itemlist}
 \item early stop: early stop can be seen as a normalization method(chapter 7 of Ref 8). The training tend to diverge in post stages. The loss curve is roughly a "U" shape thus we choose the model where loss is smallest for testing.
 \item random shuffle: We shuffle training set after training an epoch to reduce overfitting.
 \item filter out impossible observations: there is some generated observations that is not possible to happen causing the posterior of random variables doesn't sum up to 1. In our training and testing procedure, we just throw away these data.
 \end{itemlist}

\subsection{Alarm Dataset}

\begin{figure}[bh]
\centerline{\includegraphics[width=9cm]{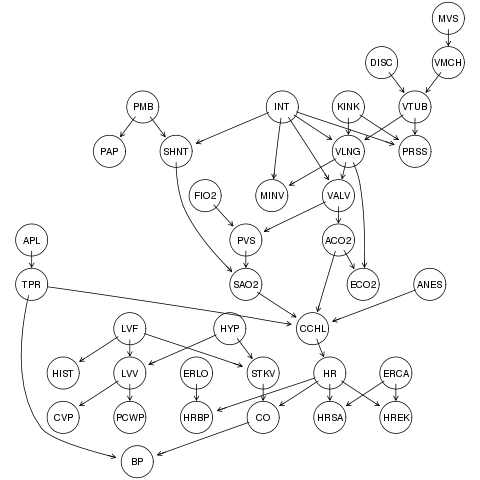}}
\vspace*{8pt}
\caption{The DAG of Alarm dataset. There are 37 nodes and 46 edges in this network. Average node size is 3.51. Number of parameter is 509.}
\end{figure}

Alarm dataset is the abbreviation for \emph{A Logical Alarm Reduction Mechanism}. It is a diagnostic application used to explore probabilistic reasoning in belief networks. We randomly draw 10,000 examples using junction tree algorithm and randomly choose 5000 examples for training DNN and 5000 for testing. The detailed network configuration is shown in table 1. Layers are densely connected. There are totally 50750 parameters. When testing using LWS algorithm, we draw 1000 samples for every observation. The number of samples is chosen by the tradeoff between running time and converging curve of LWS. Drawing more samples will lead to more accurate estimation but will be cause heavy computing burden and hard to test on multiple observations. The result is shown in table 2. Note that we only tested 320 examples from testing set for comparison because LWS is much slower than our model. The comparison of MTA metric is shown in Fig 3.
\begin{table}[th]
\tbl{The DNN parameters used in Alarm dataset.}
{\begin{tabular}{@{}ccccccc@{}} \toprule
&layer size & Hidden1 & Hidden2 & Hidden3 & Hidden4 & output layer \\ \colrule
number of nodes&105 & 100 & 150 &100 & 50 &105 \\  \botrule
\end{tabular}}
\end{table}

\begin{table}[th]
\tbl{Comparison between our model and LWS algorithm in Alarm dataset.}
{\begin{tabular}{@{}cccccc@{}} \toprule
 & time per inference & average K-L divergence & accuracy with threshold 0.1 \\
 & (/second) & &(\%)   \\ \colrule
Our model & \textbf{0.0012} &\textbf{0.0500} & \textbf{99.97} \\
LWS & 3.4482 & 1.2365 &89.61\\  \botrule
\end{tabular}}
\end{table}

\begin{figure}[bh]
\centerline{\includegraphics[width=16cm]{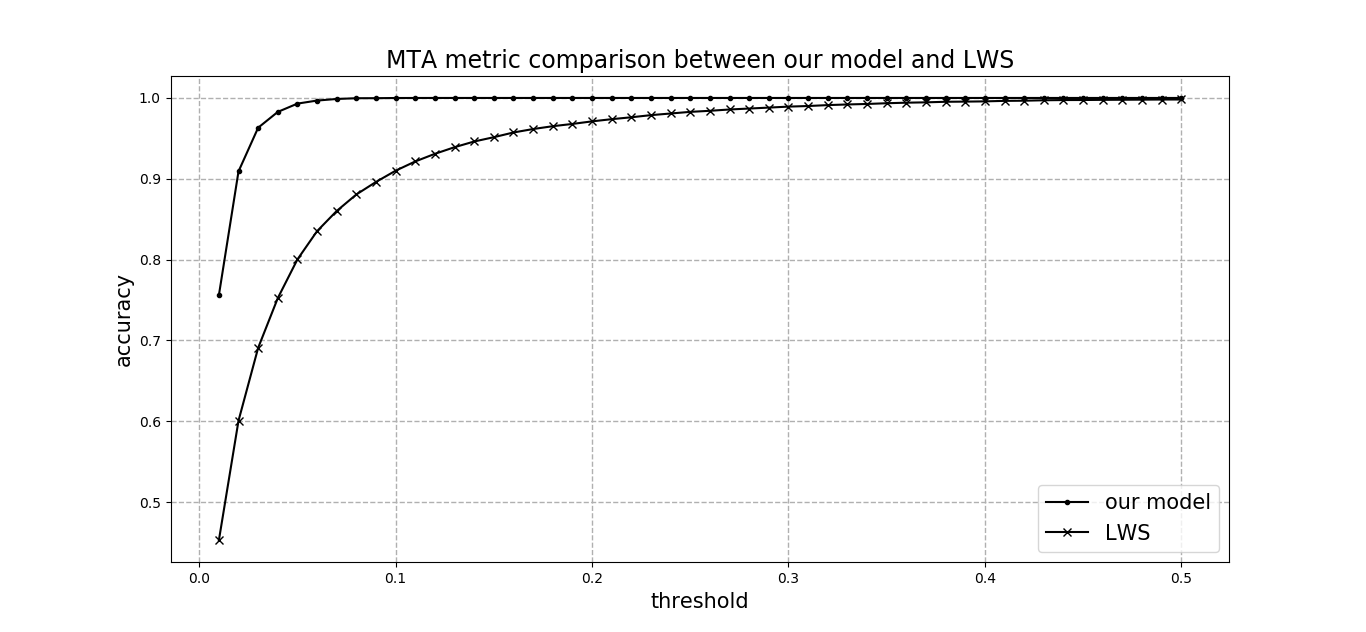}}
\vspace*{8pt}
\caption{Comparison of our model and LWS algorithm using MTA metric in Alarm dataset.}
\end{figure}

Our model is more than 2000x faster than LWS and achieve better accuracy. If accelerated with GPU, our model can run even faster. The average K-L divergence of our model is nearly 1/25 of that of LWS. And it's obvious that our model make less "big" mistake according to Fig 3.

\subsection{Insurance Dataset}

\begin{figure}[bh]
\centerline{\includegraphics[width=9cm]{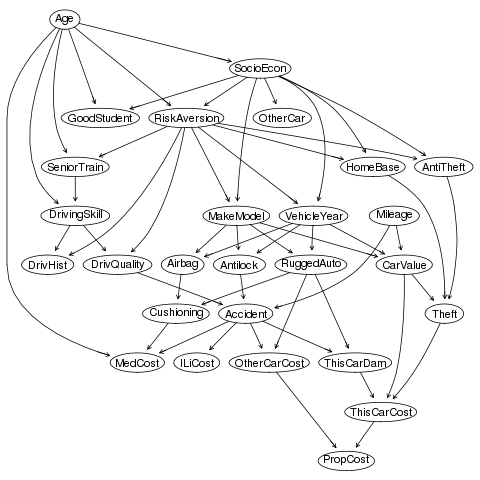}}
\vspace*{8pt}
\caption{The DAG of Insurance dataset. There are 27 nodes and 52 edges in this network. Average node size is 3.85. Number of parameter is 984.}
\end{figure}

Insurance dataset\cite{insurance} is introduced by J. Binder and his colleagues to estimate the expected claim costs for a car insurance policyholder. It is slightly bigger than Alarm dataset with nearly twice as much parameters. We use the same model configuration as Alarm dataset. The result is shown in Table 3 and Fig 5. Both Our model LWS have poorer performance in this datset comparing to Alarm dataset but our model still outperform LWS by a large margin. The two experiments show that our model has better performance than LWS in medium sized BN.

\begin{table}[th]
\tbl{Comparison between our model and LWS algorithm in Insurance dataset.}
{\begin{tabular}{@{}cccccc@{}} \toprule
 & time per inference & average K-L divergence & accuracy with threshold 0.1 \\
 & (/second) & &(\%)   \\ \colrule
Our model & \textbf{0.0001} &\textbf{0.4153} & \textbf{94.90} \\
LWS & 3.9819 & 1.6999 &82.65\\  \botrule
\end{tabular}}
\end{table}

\begin{figure}[bh]
\centerline{\includegraphics[width=16cm]{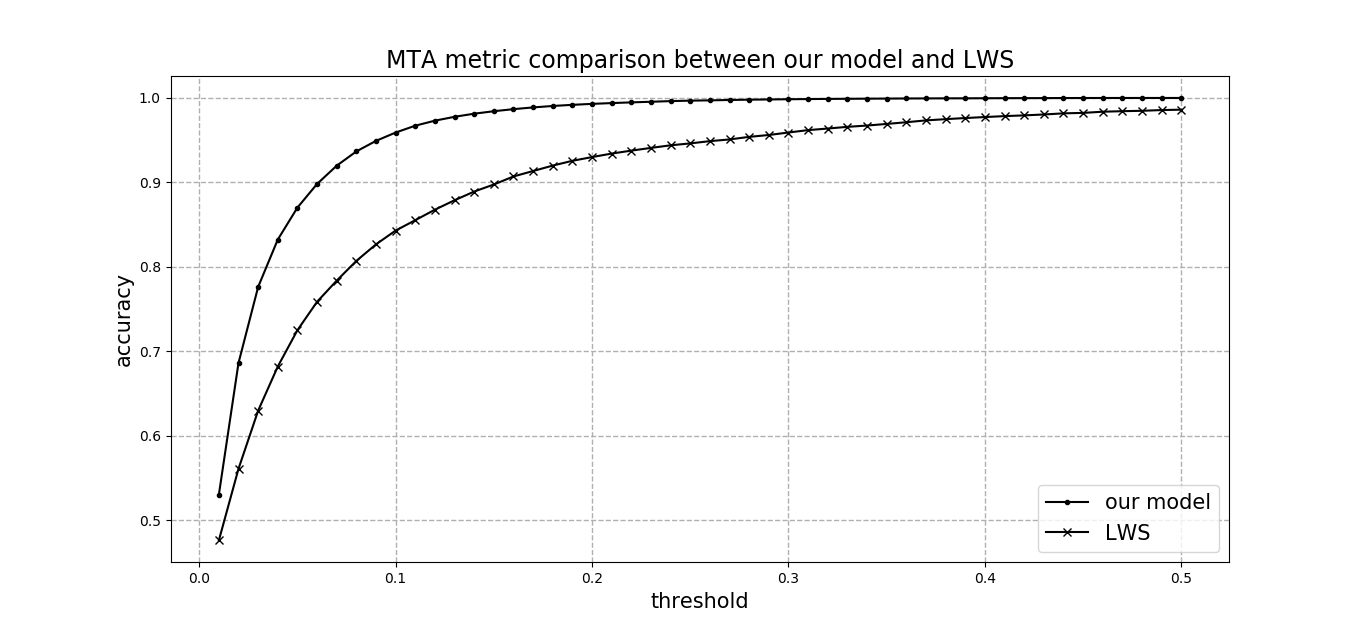}}
\vspace*{8pt}
\caption{Comparison of our model and LWS algorithm using MTA metric in Insurance dataset.}
\end{figure}

\subsection{Asia Dataset}
\begin{figure}[bh]
\centerline{\includegraphics[width=5cm]{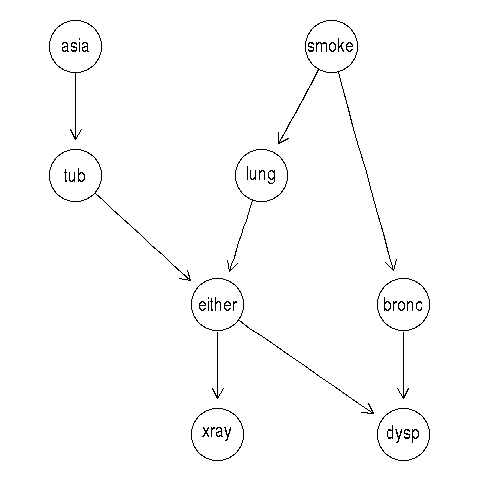}}
\vspace*{8pt}
\caption{The DAG of Asia dataset. There are 8 nodes and 8 edges in this network. Average node size is 2.50. Number of parameter is 18.}
\end{figure}
Asia \cite{asia}(sometimes called Lung Cancer dataset) is a small BN dataset containing only 8 random variables and 8 edges. Due to it's small size, we reduce our model's depth (the number of layers) and width (the number of nodes in a single layer). The DNN configuration and testing result is shown in table 4 and table 5. The MTA metric result is shown in Fig 7.

\begin{table}[th]
\tbl{The DNN parameters used in Asia dataset.}
{\begin{tabular}{@{}ccccccc@{}} \toprule
&Input layer & Hidden1 & Hidden2 & Hidden3  & output layer \\ \colrule
number of nodes&16 & 32 & 64 &32 &16 \\  \botrule
\end{tabular}}
\end{table}

\begin{table}[th]
\tbl{Comparison between our model and LWS algorithm in Asia dataset.}
{\begin{tabular}{@{}cccccc@{}} \toprule
 & time per inference & average K-L divergence & accuracy with threshold 0.1 \\
 & (/second) & &(\%)   \\ \colrule
Our model & \textbf{0.0006} &0.0669 & 95.51 \\
LWS & 3.3456 & \textbf{0.0050} &\textbf{99.29}\\  \botrule
\end{tabular}}
\end{table}

\begin{figure}[bh]
\centerline{\includegraphics[width=16cm]{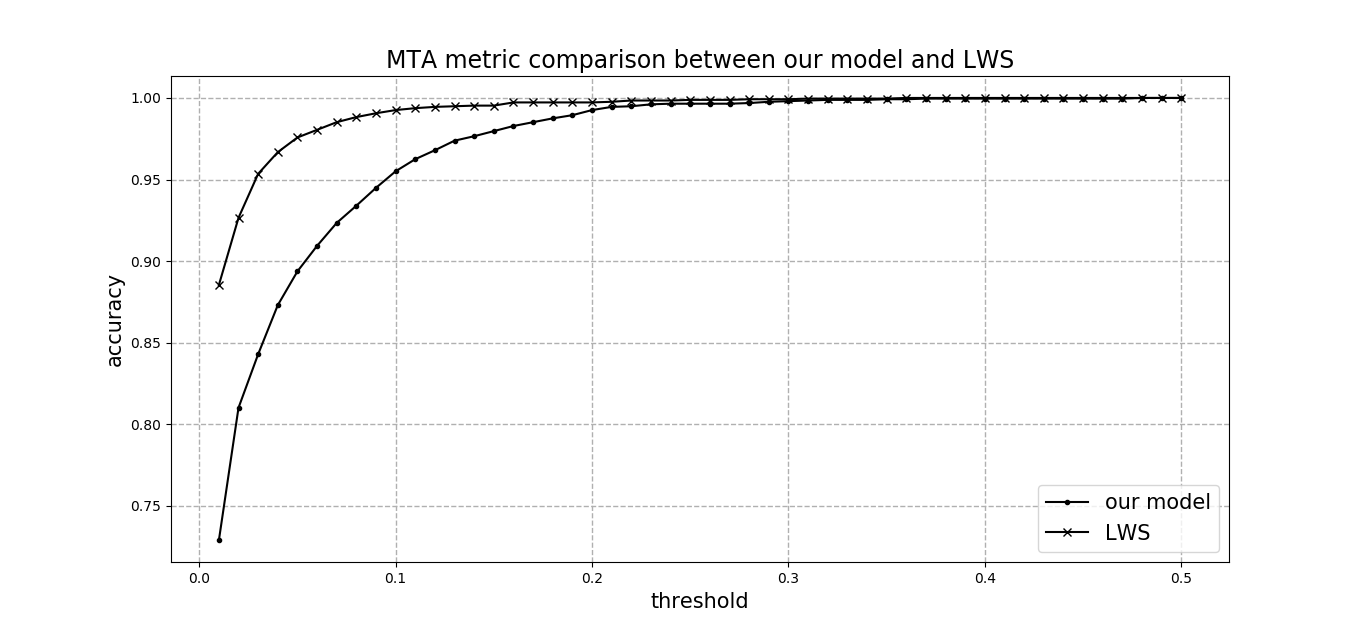}}
\vspace*{8pt}
\caption{Comparison of our model and LWS algorithm using MTA metric in Asia dataset.}
\end{figure}

Our model runs more than 5000x faster than LWS algorithm but makes more mistake. The reason might be the capacity of DNN is too large for small sized BN.

\subsection{Survey Dataset}

\begin{figure}[bh]
\centerline{\includegraphics[width=4cm]{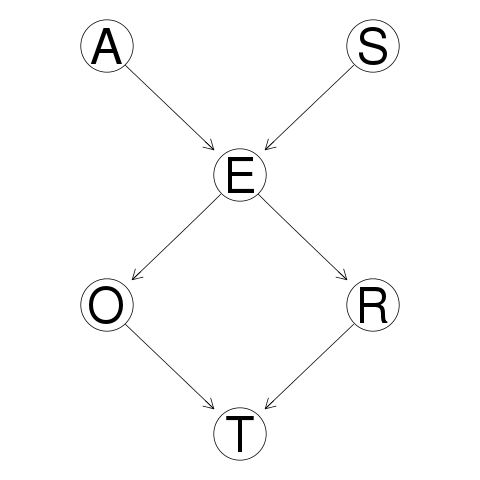}}
\vspace*{8pt}
\caption{The DAG of Survey dataset. There are 6 nodes and 6 edges in this network. Average node size is 2.67. Number of parameter is 21.}
\end{figure}
 Survey dataset\cite{survey} is another small BN dataset. We use this dataset to valid our model on small dataset. Due to it's size is similar with Asia dataset, we use the same model configuration as Asia dataset. The result is shown in table 6 and Fig 9.Our model run more than $10^4$x faster than LWS algorithm with slightly accuracy loss. The result of Asia dataset and Survey dataset indicate our model is weaker in modeling small BN but can still get reasonable result.

\begin{table}[th]
\tbl{Comparison between our model and LWS algorithm in Survey dataset.}
{\begin{tabular}{@{}cccccc@{}} \toprule
 & time per inference & average K-L divergence & accuracy with threshold 0.1 \\
 & (/second) & &(\%)   \\ \colrule
Our model & \textbf{0.0003} &0.0058 & 99.96 \\
LWS & 3.0740 & \textbf{0.0013} &\textbf{100.00}\\  \botrule
\end{tabular}}
\end{table}

\begin{figure}[bh]
\centerline{\includegraphics[width=16cm]{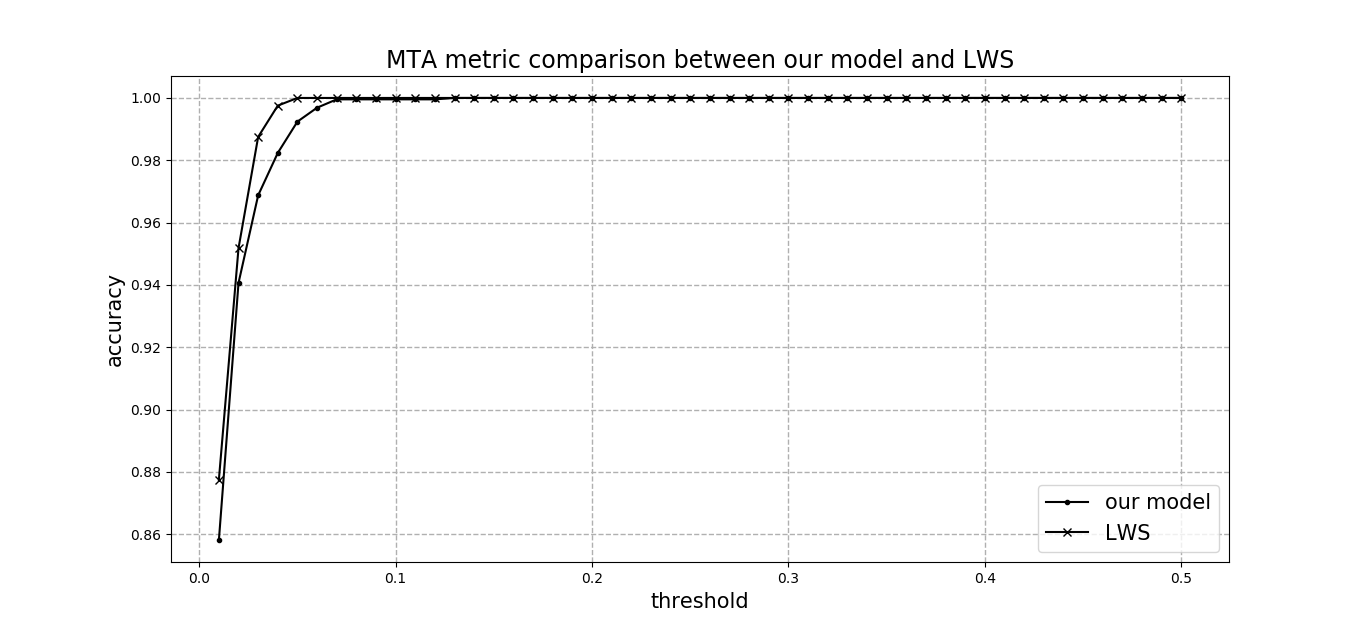}}
\vspace*{8pt}
\caption{Comparison of our model and LWS algorithm using MTA metric in Survey dataset.}
\end{figure}

\section{How many examples needed}
\begin{figure}[bh]
\centerline{\includegraphics[width=16cm]{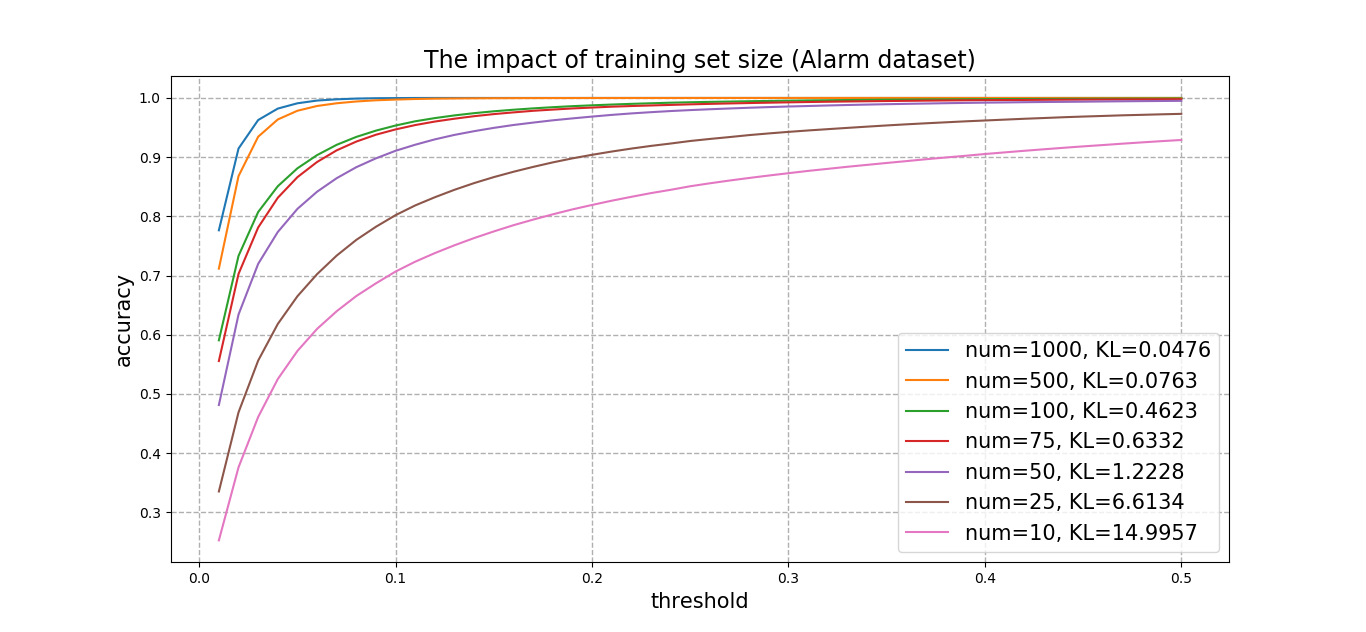}}
\vspace*{8pt}
\caption{The MTA curve of different models trained with different number of examples.}
\end{figure}

We did more experiments on Alarm dataset by changing the number of examples in training set and inspect its inspect on accuracy of our model. The result is shown in Fig 10. Our model can get almost as good result when the size of training set drop from 5000 to 500. After further decreasing the size of training set, the accuracy of our model drops remarkably. Another detail about our model is that training accuracy and testing accuracy are almost the same which means there is no overfitting in our training process.

The detailed comparison between our method and LWS is illustrated in Table 7. In general, the advantages of our method include faster speed and better accuracy in medium sized BN. The disadvantages are that we cannot gain better accuracy with more training examples once our model saturate and we can't mathematically prove the convergence property of our model. What's more, our model rely on other inference algorithms to generate training examples.

\begin{table}[th]
\tbl{Detailed comparison between our model and LWS algorithm}
{\begin{tabular}{@{}llll@{}} \toprule
  &Our method & LWS \\ \colrule
Running time & Fast, polynomial time complexity. & Slow, exponential time complexity. \\
 Accuray & Better accuracy in medium sized BN, &Better accuracy with more samples,inferior\\
  &but has accuracy limitation. & accuracy when running time is limited.\\
 Scalability& Poor scalability, we have to train a& Good scalability.\\
&   new network for a new BN.&\\
 Convergence & Cannot be proved mathematically. & Proved convergence.\\ \botrule
\end{tabular}}
\end{table}

\section{Future works}

Our model currently depend on exact inference algorithm to generate training examples. For those BNs where exact inference is impossible (due to exponential  time complexity), our model cannot be trained. Future works might enable DNN learn directly from raw data (samples draw from joint probability distribution from BN).

\section{Conclusion}

Our method is much faster than traditional approximate inference algorithm LWS and achieve better accuracy on medium sized BN. Our method has the same defect as traditional inference algorithms\cite{realtimeinference}. Both may fail to converge to reasonably good result on some dataset.



\end{document}